\documentclass[conference]{IEEEtran}
\IEEEoverridecommandlockouts
\usepackage{comment}
\usepackage{graphicx}
\usepackage{multirow}
\usepackage{booktabs}
\usepackage{cite}
\usepackage{amsmath,amssymb,amsfonts}
\usepackage{algorithmic}
\usepackage{graphicx}
\usepackage{textcomp}
\usepackage{xcolor}
\usepackage{url}
\def\BibTeX{{\rm B\kern-.05em{\sc i\kern-.025em b}\kern-.08em
    T\kern-.1667em\lower.7ex\hbox{E}\kern-.125emX}}
    
\begin{document}

\title{TRACE: Temporal Relationship-Aware Conversational Entrainment Detection in Dyadic Speech}
 \author{
 \IEEEauthorblockN{
 Sathvik Manikantan Napa Ugandhar\textsuperscript{1,*,$\dagger$},
 Hao Zhang\textsuperscript{1,*,$\dagger$},
 Alison Gunzler\textsuperscript{1},
 Yuzhe Wang\textsuperscript{2}
 }
 \IEEEauthorblockN{
 Thomas Thebaud\textsuperscript{2},
 Georgi Tinchev\textsuperscript{3},
 Venkatesh Ravichandran\textsuperscript{4},
 Laureano Moro-Velázquez\textsuperscript{2,$\dagger$}
 }
 \IEEEauthorblockA{
 \textsuperscript{1}Department of Computer Science, Johns Hopkins University, USA\\
 \textsuperscript{2}Department of Electrical and Computer Engineering, Johns Hopkins University, USA\\
 \textsuperscript{3}Amazon Research, UK\\
 \textsuperscript{4}Amazon, USA\\
 snapaug1@jh.edu, hzhan276@jh.edu, laureano@jhu.edu
 }
 \thanks{
 * Equal contribution. \quad
 $\dagger$ Corresponding author.
 }
 }

\maketitle

\begin{abstract}
With the proliferation of speech AI agents, understanding emotional entrainment in conversational interaction has become increasingly important. Emotional entrainment is shaped by social relationships and conversational context, influencing affective coordination over time. In this study, we introduce DyadEE, a dataset for emotional entrainment detection in dyadic speech interactions, containing both emotionally entrained conversations and synthetic interactions where entrainment is disrupted through emotion resynthesis. We further propose TRACE, a window-level framework that models dyadic interaction as ordered sequences of acoustic embeddings derived from emotion fine-tuned Whisper representations, treating each sample as an interaction trace rather than pooled utterances. Experimental results on DyadEE show that incorporating conversational context and relationship information improves emotional entrainment detection, with TRACE achieving 93.47\% accuracy.
\end{abstract}


\section{Introduction}
Entrainment is the natural tendency of two participants in a conversation to align with each other over the course of an interaction. 
Dyadic speech interaction manifests across multiple dimensions, including speech rate, lexical choice, prosody, and affective adaptation \cite{wynn2022rhythm,kruyt2023measuring}.
Among these, \textit{emotional entrainment}, the mutual adaptation of affective states between speakers during interaction, is strongly shaped by the relationship between speakers and the conversational context \cite{kejriwal2022relationship}. 
In natural conversations, emotional alignment is not limited to mirroring:
partners converge affectively in affiliative contexts, while complementary or regulatory responses are more natural in conflict, negotiation, or support-seeking scenarios \cite{zee2023physiological}. 
Emotional entrainment is therefore best understood as a relationship-conditioned and context-conditioned process rather than a single context-free global similarity score \cite{bevnuvs2014social}.

Understanding emotional entrainment has direct practical value for conversational speech AI. In speech-to-speech agent systems, a response style that is appropriate in peer interaction may be inappropriate in a clinician-patient setting. Models deployed in emotionally sensitive domains such as companionship, mental health support, or professional assistance must therefore adapt their prosodic and affective behavior to both their social role and the evolving conversational context \cite{phang2025investigating, thakkar2024artificial}. Evaluations that ignore these factors can produce misleading conclusions, rewarding globally pleasant responses that violate role-specific norms and masking failures that emerge only under particular interaction settings \cite{sun2024unpacking}. This motivates benchmarks that assess emotional entrainment under explicit contextual and relational constraints.

Prior work has explored related aspects of affective coordination in conversation, including empathetic alignment in NLP, emotional synchrony in dyadic interaction, and acoustic--prosodic entrainment in speech \cite{levitan2011measuring,yang2024modeling}. 
These lines of work suggest that alignment is shaped not only by lexical content, but also by temporal interaction structure and broader conversational conditions \cite{mcneill2024autoregressive, herbuela2025spatiotemporal}.

Prior modeling approaches range from lightweight speech emotion recognition baselines such as Emotion MLP \cite{joy2020speech} to long-range dyadic interaction models such as DyadFormer \cite{curto2021dyadformer}, but neither is designed to evaluate emotional entrainment under explicit conversational context and social relationship constraints. This gap motivates our study.

Direct measurement of emotional entrainment would require dense, continuous annotation of affective states across both speakers throughout a conversation, an annotation burden that restricts prior work to small, controlled corpora. We instead operationalize entrainment detection through \textit{controlled disruption}: non-entrained dyads are constructed by specifically severing inter-speaker affective coordination via emotion-contradicting resynthesis (which inverts one speaker's affective trajectory while preserving lexical content), while maintaining surface acoustic realism. These disruptions are targeted interventions, not arbitrary mismatches: each is designed to break the mutual adaptation mechanism that defines entrainment, not merely to produce audio that sounds wrong.

We are explicit that detection accuracy does not directly quantify the degree of entrainment in a dyad; it serves instead as a diagnostic of which relational and contextual signals drive affective coordination in speech. This framing motivates two research questions: \textbf{(RQ1)} Can disrupted affective coordination be reliably detected from speech under diverse relational and contextual conditions? \textbf{(RQ2)} Do relationship type and conversational context provide complementary, architecture-sensitive signals for this detection?  To investigate these questions, we introduce DyadEE, built on the Seamless Interaction corpus \cite{agrawal2025seamless}, which pairs natural dyads with controlled disruptions across 7 relationship categories and 20 conversational contexts, with voice-converted entrained dyads to prevent artifact-based shortcuts. We model each dyad as a temporal sequence of window-level acoustic embeddings from an emotion-finetuned Whisper encoder and show that relationship and context conditioning each provide distinct gains that are architecture-dependent and largely complementary.

\begin{figure*}[t]
  \centering
  \includegraphics[width=1\textwidth]{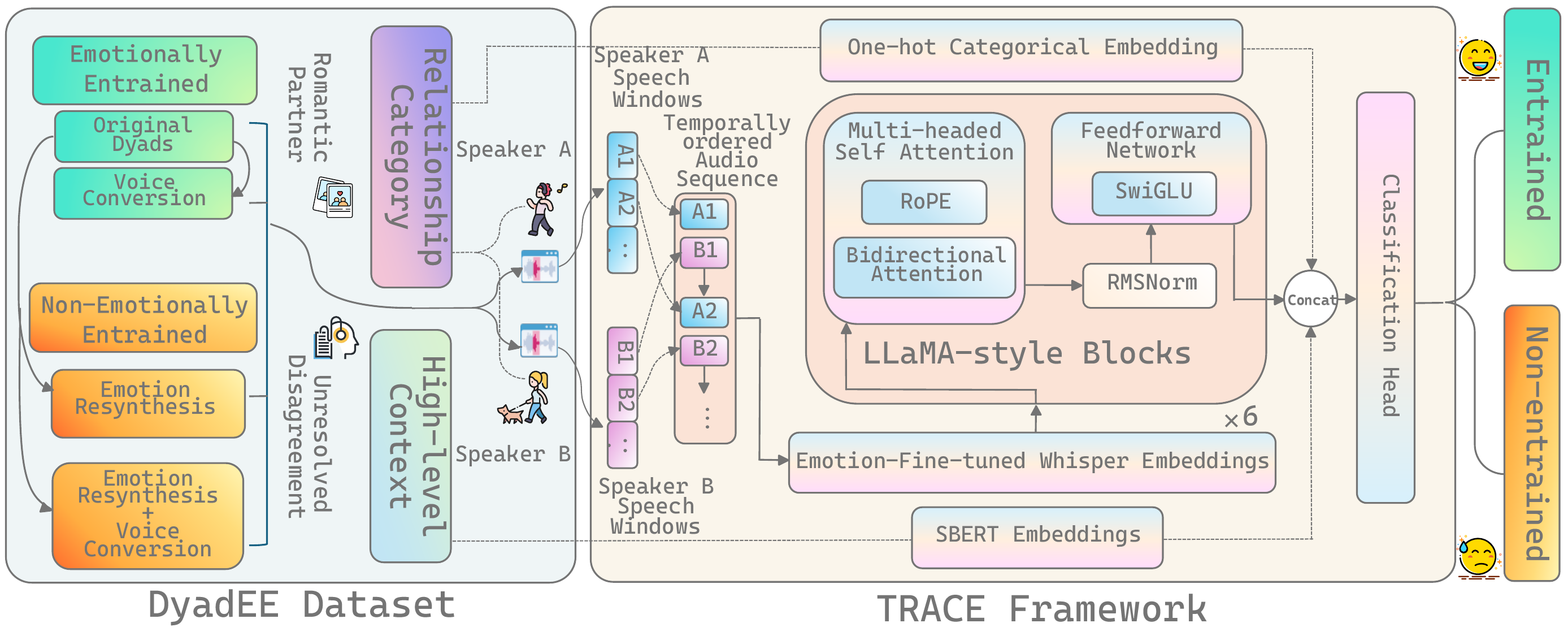}
  \caption{TRACE pipeline.Left: DyadEE dataset construction with emotion-entrained dyads (original conversations and their voice-converted variants, denoised via MossFormer2 as a preprocessing step) and disrupted dyads generated via emotion-contradicting resynthesis, each annotated with conversational context and relationship category. Right: TRACE models dyadic interaction using window-level embeddings from an emotion-fine-tuned Whisper encoder for emotional entrainment classification.}
  \vspace{-16pt}
  \label{fig:TRACE}
  
\end{figure*}


Our contributions are: (i) Temporal Relationship-Aware Conversational Entrainment Discriminator (TRACE), which addresses \textbf{RQ2} by comparing two integration architectures (late fusion and cross-attention) for incorporating relationship and context into temporal dyadic sequence modeling; and (ii) Dyadic Emotional Entrainment Dataset (DyadEE), which addresses \textbf{RQ1} through principled disruption strategies and voice-conversion augmentation to prevent acoustic shortcuts. We release both the dataset and codebase to support future research.\footnote{\url{https://github.com/anonymoususer276/TRACE}}

\section{Dyadic Emotional Entrainment Dataset}
We introduce DyadEE, a dataset for emotional entrainment detection in dyadic speech interaction. Figure \ref{fig:TRACE} (left) summarizes its construction from entrained dyads, including augmented entrained variants obtained via voice conversion, and emotion resynthesis, as explained in the next subsections. All original recordings are additionally denoised via MossFormer2 speech enhancement as a preprocessing step prior to voice conversion and resynthesis. Consequently, each dyad will have a label indicating if it is emotionally entrained (positive) or non-emotion entrained (negative). Table \ref{tab:dataset_composition} reports the data types, number of dyads, and total duration.
\subsection{Selection of Natural Dyadic Interactions}
We build DyadEE on top of the Seamless Interaction dataset \cite{agrawal2025seamless}, a large-scale collection of entrained dyadic conversations recorded under diverse speaker prompts designed to elicit rich social and emotional interactions. 
The dataset provides relationship metadata and specific contextual prompts that make it well-suited for studying emotional entrainment in conversational speech. 

In our curated dataset, the relationship distribution includes 7 categories and their counts in parentheses: friends (5,484), family-generic (931), romantic partners (819), coworkers (756),  classmates (320), siblings (92) and Parent-Child (32). 

\begin{table}[t]
\centering
\caption{DyadEE dataset composition across emotionally entrained (EE) and non-emotionally entrained (NE) dyadic speech, with various controlled perturbations}
\footnotesize
\setlength{\tabcolsep}{1.7pt}
\renewcommand{\arraystretch}{1.08}
\vspace{-2mm}
\resizebox{\linewidth}{!}{  

\begin{tabular}{l c c c c}
\toprule
\textbf{Data Type} & \textbf{Total} & \textbf{Train} & \textbf{Test} & \textbf{Duration (h)}\\
\midrule
EE + Original Dyad               & 2{,}125 & 1{,}295 & 830 & 230.1 \\
EE + Voice Conversion            & 2{,}113 & 1{,}288 & 825 & 232.0 \\
NE: Emotion Resynthesis          & 2{,}125 & 1{,}295 & 830 & 317.4 \\
NE: Emotion Resynthesis + Voice Conversion     & 2{,}071 & 1{,}253 & 818 & 290.3 \\
\midrule
\textbf{Total} & \textbf{8{,}434} & \textbf{5{,}131} & \textbf{3{,}303} & \textbf{1{,}069.8} \\
\bottomrule
\end{tabular}}
\label{tab:dataset_composition}
\vspace{-10pt}
\end{table}

To focus on emotionally informative interactions, we abstract the original fine-grained
speaker prompts from the Seamless Interaction subsample into 20 general context categories. The Seamless corpus contains 798 unique speaker prompts; we prompted OpenAI GPT-5~\cite{singh2025openai} to assign each prompt a thematic label capturing its emotional situation. The categorization criterion prioritized emotional salience; categories were defined by the affective challenge or relational dynamic at stake (e.g., delivering difficult news, navigating conflict, seeking support) rather than by topic domain alone. We then manually reviewed all 798 prompt-to-category mappings, consolidated near-duplicate categories, and corrected misassigned prompts, yielding the final set of 20 categories. Their DyadEE dyad counts are: isolation and support needs (653), unresolved disagreements (618), embarrassing stories (586), mistakes and apologies (581), regret and apologies (565), failure and lessons (560), support experiences (506), trust and decision-making (448), consideration and misunderstandings (436), relationship withdrawal (399), leadership and decision-making (398), self-defense events (392), delivering bad news (378), overcoming challenges (378), challenges and support (363), communication and feelings (292), habits and change (285), past compromise (257), difficult work relationships (175), and missed chance to speak up (164). These contexts are used to curate emotionally entrained dyads and to construct context-controlled disrupted interactions.

\subsection{Dyad Expansion Strategies}

We expand DyadEE through controlled construction of non-entrained dyads and augmentation of entrained dyads.
    
\noindent\textbf{Non-Emotionally Entrained (NE): Emotion Resynthesis.} For each conversation, one speaker is randomly selected for resynthesis while the other's original recording is kept unchanged. We run VoxProfile~\cite{feng2025vox} on the selected speaker's audio to estimate their dominant emotion from nine categories: \textit{Neutral, Happy, Sad, Angry, Contempt, Fear, Disgust, Surprise}, and \textit{Other}. We then retrieve the speaker's original transcript directly from the Seamless Interaction dataset and resynthesize their channel using an expressive Text-To-Speech (TTS) system, EmotiVoice~\cite{netease_youdao_emotivoice}, in a deliberately contradicting emotional style (e.g., \textit{Happy}$\rightarrow$\textit{Sad}, \textit{Angry}$\rightarrow$\textit{Neutral}). This one-sided manipulation disrupts emotional entrainment while preserving conversational content, yielding non-entrained dyads that retain one emotion entrained human channel.

\noindent\textbf{Emotionally Entrained (EE): Voice Conversion.}
A potential shortcut for an entrainment classifier is to treat synthesis artifacts or vocoder traces, as a proxy for non-entrainment, since NE: Emotion Resynthesis dyads contain a TTS-resynthesized channel. In this study, we consider that TTS and neural-based voice conversion (VC) produce similar acoustic artifacts but with opposite entrainment labels: TTS replaces the acoustic realization and prosody with a synthetic rendering, breaking entrainment, whereas VC re-synthesizes speech modifying speaker identity onto the original recording, fully preserving the source speaker's prosody, rhythm, and emotional trajectory. By including VC-augmented entrained dyads alongside TTS-based non-entrained dyads, the training set contains synthesized examples on both sides of the label boundary, forcing the model to attend to emotional coordination rather than acoustic authenticity.

Concretely, we apply SeedVC~\cite{li2024seedvc} to each entrained dyad via cross-speaker self-conversion: speaker $A$'s recording is converted to the voice of speaker $B$ (using $B$'s channel as the voice reference), and speaker $B$'s recording is converted to the voice of speaker $A$ (using $A$'s channel as the reference). This produces a fully synthetic dyad that preserves the original interaction dynamics, timing, turn-taking, and emotional trajectory, with both speaker identities swapped. We additionally apply MossFormer2~\cite{zhao2024mossformer2} speech enhancement to all original recordings prior to processing to improve signal quality in noisy conditions. Finally, applying SeedVC to NE: Emotion Resynthesis dyads yields a fourth data
type (NE: Emotion Resynthesis + VC), in which both speaker identities are
additionally converted, producing a fully synthetic non-entrained dyad that
further decouples the non-entrainment signal from any single speaker's voice
characteristics.

\section{TRACE Framework}
We propose TRACE, an emotional entrainment classifier that leverages a temporally ordered sequence of alternating speaker windows and explicitly conditions prediction on acoustic, contextual, and relationship information to determine whether a dyadic interaction is emotionally entrained or non-entrained. Figure~\ref{fig:TRACE} (right) illustrates the TRACE framework.
\subsection{Dyadic Representation and Input Features}

\noindent\textbf{Temporal Dyadic Representation:} Given a dyad with two speakers $A$ and $B$ recorded in separate channels, we extract fixed-length $t$-second windows independently from each channel, yielding $N$ windows per speaker. The two sequences are then interleaved into a single temporal sequence: $A_{1} \rightarrow B_{1} \rightarrow A_{2} \rightarrow B_{2} \rightarrow \cdots
\rightarrow A_{N} \rightarrow B_{N}$, where $A_{n}$ and $B_{n}$ denote the $n$-th window from each speaker's channel and correspond to the same temporal position in the conversation, not concurrent overlap. If the two channels differ in length, the shorter is zero-padded; conversations exceeding $N_{\max}{=}256$ windows per speaker are truncated to the first $N_{\max}$ windows (covering up to $N_{\max}{\times}t$ seconds). This representation offers advantages over turn-level modeling: (i) it requires no turn detection or forced alignment, which are error-prone under overlapping speech; (ii) fixed-length windows provide a regular temporal grid, making positional encoding stable and meaningful across conversations of varying length; and (iii) it captures within-turn affective dynamics that may be smoothed over by utterance- or turn-level pooling.

\noindent\textbf{Acoustic Emotion Embeddings:} For each speech window, we extract a dense acoustic embedding from the final hidden layer of an emotion-finetuned Whisper-large-v3 encoder \cite{timmel2025fine} provided by the VoxProfile suite~\cite{feng2025vox}\footnote{Pretrained model accessed on January 8th, 2026}. The encoder is trained for valence, arousal, and dominance prediction, enabling it to capture affect-relevant prosodic and acoustic cues such as intonation, intensity, and spectral variation over short temporal segments.
The resulting window-level embeddings are arranged according to the temporal dyadic sequence and serve as the primary acoustic input to TRACE. This allows the model to track how local affective patterns evolve across alternating speaker windows rather than relying on a single utterance-level summary.


\noindent\textbf{Context and Relationship Features:} In addition to the acoustic emotion embeddings, TRACE takes two dyad-level auxiliary inputs: conversational context and speaker-pair relationship. Both are encoded as 384-dimensional embeddings using sentence-BERT (all-MiniLM-L6-v2)~\cite{reimers2019sentence}: the conversational context from the speaker prompt text, and the relationship from its category label rendered as a short text string (e.g., \textit{``friends''}, \textit{``coworkers''}). Unlike the window-level acoustic sequence, these signals are fixed for a given dyad and capture global social and situational information about the interaction.

\subsection{Emotional Entrainment Modeling}
\noindent\textbf{Emotional Interaction Modeling:} Given the temporal sequence of window-level acoustic emotion embeddings, TRACE models dyadic interaction with a lightweight stack of six LLaMA-style blocks \cite{touvron2023llama}. Each block adopts RoPE positional encoding, RMSNorm, and feed-forward layers with SwiGLU activation, while replacing the original causal attention with bidirectional self-attention to capture dependencies across both preceding and following speaker windows. This design is better suited to dyadic sequence classification, where emotional entrainment depends on interaction patterns distributed across the full conversation rather than left-to-right generation alone. Through these stacked blocks, TRACE transforms local acoustic emotion cues into higher-level representations that encode interaction dynamics and emotional coordination across the dyad. 

\noindent\textbf{Feature Integration and Classification:} To incorporate conversational context and social relationship into entrainment detection, we encode both signals as fixed 384-dimensional embeddings using sentence-BERT (all-MiniLM-L6-v2)~\cite{reimers2019sentence} and project them to 128 dimensions each. We evaluate two conditioning strategies. \textit{TRACE-LF (Late Fusion)} applies a masked mean pool over the audio token sequence, concatenates the result with the projected context and relationship embeddings, and feeds the combined representation into a two-layer MLP binary classifier. \textit{TRACE-CE (Cross-Encoder)} instead constructs a single auxiliary token from the projected context and relationship embeddings and passes it through two cross-attention layers (8 heads, GELU feed-forward, dropout\,=\,0.1) in which the auxiliary token queries the full audio token sequence; the resulting fused token is fed to the classification head. Both variants are trained with AdamW ($\text{lr}{=}10^{-4}$), 2-epoch linear warmup followed by cosine decay over 25 epochs, Gaussian noise ($\sigma{=}0.15$) and 30\% token masking on speech embeddings during training only; TRACE-LF uses batch size 512 and TRACE-CE batch size 256.


\section{Experiments}


\label{sec:exp_setup}
Our experiments include two parts: a performance comparison of our system with existing baselines and an ablation study.

We compare TRACE with two baselines: Emotion MLP~\cite{joy2020speech}, a lightweight speech emotion recognition model, and DyadFormer~\cite{curto2021dyadformer}, a Transformer for long-range dyadic interaction modeling. All models use the same acoustic feature source: window-level embeddings from the emotion-finetuned Whisper encoder and the same SBERT-encoded context and relationship embeddings where applicable. The models differ in how they process the acoustic sequence: Emotion MLP aggregates speaker embeddings via mean pooling, discarding temporal order entirely; DyadFormer applies its own dyadic sequence model over the windows; and TRACE uses the alternating-speaker temporal representation described above. This setup ensures that performance differences reflect the temporal modeling architecture rather than the choice of features. We report accuracy, ROC-AUC, and macro F1.


Our ablation study evaluates four input configurations as follows: \textbf{Speech only} (Sp), \textbf{Speech$+$Context} (Sp$+$Ctx), \textbf{Speech$+$Relationship} (Sp$+$Rel), and \textbf{Speech$+$Context$+$Relationship} (Sp$+$Ctx$+$Rel), and further analyzes model behavior across relationship and context categories.

All models share the same data split and training setup. We partition DyadEE into 5{,}131 training dyads and 3{,}303 held-out test dyads (approximately 60:40), with 10\% of the training set reserved as a validation set (513 dyads) for checkpoint selection. The split is speaker-disjoint, preventing identity-based leakage. All models are optimized with cross-entropy loss on a single NVIDIA H200 GPU.

\subsection{Evaluation Data}


\noindent\textbf{Data leakage validation.}\hspace{0.4em} We verify that DyadEE is free from the main sources of cross-split leakage. \textit{Conversation-level integrity:} the train/test split is performed at the speaker-pair level: no two speakers who form a pair in training appear together in the test set. Since all four data types of a conversation (EE+Original Dyad, EE+Voice Conversion, NE: Emotion Resynthesis, and NE: Emotion Resynthesis+VC) share the same two speaker identities, speaker-pair disjointness guarantees that no variant of any conversation crosses the split boundary. \textit{Voice conversion targets:} EE+Voice Conversion dyads are produced via within-conversation self-conversion (speaker $A$ converted to speaker $B$'s voice and vice versa within the same dyad), so no external speaker voice acts as a cross-split signal.

\noindent\textbf{Metadata shortcuts:} To verify that emotion entrainment cannot be inferred from context and relationship labels alone, we train classifiers using only these two features. Logistic regression achieves 50.11\% accuracy and a depth-10 decision tree 50.02\%, both at random chance. This is by construction: every (context, relationship) pair contains an equal number of EE and NE dyads, as each EE+Original Dyad has a matched NE: Emotion Resynthesis counterpart under identical social conditions. One speaker identity (out of 4,250) appears in a small number of different conversations across splits (12 rows total); its effect is negligible and further mitigated by \textit{EE+Voice Conversion} augmentation, which reduces model reliance on speaker identity cues.


\subsection{Human Evaluation}
\label{sec:human_eval}

To assess whether the binary entrainment labels in DyadEE correspond to
perceptible differences in emotional coordination, we conducted a listening study on a stratified sample of 80 dyads drawn from the DyadEE test set, covering all four data types: EE+Original Dyad (2\%), EE+Voice Conversion (2\%), NE: Emotion Resynthesis (2\%), and NE: Emotion Resynthesis+Voice Conversion (2\%).
Ten evaluators each rated 22 clips, with each clip receiving an average of 2.75 independent ratings.
Raters listened to the full two-speaker mixed audio and scored overall conversational emotion entrainment (i.e. How well do Speaker A and Speaker B's emotions match or change naturally across the clip?) on a 5-point Likert scale (1\,=\,highly  unmatched (unnatural), 5\,=\,highly matched (natural)), without access to labels or system information.

Table~\ref{tab:human_eval} reports the mean emotion entrainment score per data type, averaged across raters at the clip level.
Emotionally entrained dyads (EE+Original Dyad and EE+Voice Conversion) received substantially higher scores ($\mu = 4.49$, $\sigma = 0.47$) than non-entrained dyads produced via emotion-contradicting resynthesis ($\mu = 2.68$, $\sigma = 0.73$), a difference that is highly significant (Mann--Whitney $U$, $p < 0.001$).
Notably, voice-converted entrained dyads (EE+Voice Conversion, $\mu = 4.25$) remain well above all non-entrained conditions, confirming that voice conversion does not undermine perceived emotion entrainment of the interaction.
The NE: Emotion Resynthesis+Voice Conversion condition scores higher than NE: Emotion Resynthesis alone ($3.03$ vs.\ $2.21$), consistent with voice conversion partially masking TTS artifacts while leaving the underlying emotional mismatch intact. Raters' open-ended responses further corroborate these patterns: 36\% of low-emotion-entrainment judgements on non-entrained clips cited unnatural speech delivery and 30\% cited emotionally flat or exaggerated affect, whereas 87\% of entrained-clip ratings required  no identified deficiency (scored 4--5). These results confirm that the automatic entrainment labels in DyadEE correspond to a perceptible distinction in dyadic emotional coordination, supporting the use of controlled disruption as a proxy task for studying entrainment-related interaction dynamics.

\begin{table}[t]
\centering

\setlength{\tabcolsep}{10pt}
\caption{Human evaluation emotion entrainment scores (1--5) per data type. EE\,=\,Emotionally Entrained, NE\,=\,Non-Emotionally Entrained.}
\begin{tabular}{l c c c}
\toprule
\textbf{Data Type} & \textbf{Label} & \textbf{Mean $\pm$ Std} \\
\midrule
EE+Original Dyad                      & EE & $4.74 \pm 0.24$ \\
EE+Voice Conversion                   & EE & $4.25 \pm 0.51$ \\
NE: Emotion Resynthesis+Voice Conv.   & NE & $3.03 \pm 0.61$ \\
NE: Emotion Resynthesis               & NE & $2.21 \pm 0.61$ \\
\midrule
EE (all)                              &    & $4.49 \pm 0.47$ \\
NE (all)                              &    & $2.68 \pm 0.73$ \\
\bottomrule
\end{tabular}
\label{tab:human_eval}
\vspace{-8pt}
\end{table}

\subsection{Comparative Results}
Table~\ref{tab:results} compares the performance of different models across various feature combinations using accuracy, ROC-AUC, and macro F1 metrics on the test set. The results show complementary roles for relationship and context conditioning across different architectures. TRACE-CE achieves the highest speech-only accuracy (86.82\%) and benefits strongly from 
relationship conditioning (93.47\%, $+$6.65\,pp), the largest absolute gain observed across any model and feature configuration. Context conditioning alone, however, reduces TRACE-CE accuracy to 83.92\% ($-$2.90\,pp), and joint conditioning likewise yields 83.98\% ($-$2.84\,pp), indicating that relationship is the primary conditioning signal for this architecture. TRACE-LF shows the complementary pattern: it gains most from context conditioning 
(83.59\%\,$\to$\,88.71\%, $+$5.12\,pp) while relationship conditioning provides no additional  benefit ($-$0.71\,pp). DyadFormer \cite{curto2021dyadformer}, a transformer-based sequence model, achieves competitive  speech-only accuracy (83.89\%) with a marginal context gain ($+$0.61\,pp), but degrades  substantially under relationship conditioning ($-$10.50\,pp to 73.39\%), indicating that its sequential encoder is not suited to cross-modal conditioning via social-role embeddings. Emotion MLP~\cite{joy2020speech}, which discards temporal order entirely through mean pooling, performs near chance on speech alone (51.07\%) and with context (50.24\%), but recovers substantially with relationship conditioning (80.29\%), suggesting that social role provides a coarse but recoverable prior when temporal structure is absent. Across models, relationship conditioning is the more consistently beneficial auxiliary signal, while context conditioning benefits are architecture-dependent.

\begin{table}[t]
\centering
\footnotesize
\setlength{\tabcolsep}{7.5pt}
\renewcommand{\arraystretch}{1.2}
\caption{Accuracy (\%), ROC-AUC, and macro F1 on the test set. Sp\,=\,Speech Only, Ctx\,=\,Context, Rel\,=\,Relationship. Best value per metric within each feature group is in \textbf{bold}.}
\begin{tabular}{l l c c c}
\hline
\textbf{Features} & \textbf{Model} & \textbf{Acc.} & \textbf{ROC-AUC} & \textbf{F1} \\
\hline
\multirow{4}{*}{Sp}
  & Emotion MLP~\cite{joy2020speech} & 51.07 & 0.685 & 0.356 \\
  & DyadFormer~\cite{curto2021dyadformer} & 83.89 & 0.924 & 0.839 \\
  & TRACE-LF & 83.59 & 0.909 & 0.836 \\
  & \textbf{TRACE-CE} & \textbf{86.82} & \textbf{0.999} & \textbf{0.866} \\
\hline
\multirow{4}{*}{Sp+Ctx}
  & Emotion MLP~\cite{joy2020speech} & 50.24 & 0.526 & 0.334 \\
  & DyadFormer~\cite{curto2021dyadformer} & 84.50 & 0.930 & 0.845 \\
  & \textbf{TRACE-LF} & \textbf{88.71} & 0.996 & \textbf{0.886} \\
  & TRACE-CE & 83.92 & \textbf{0.998} & 0.835 \\
\hline
\multirow{4}{*}{Sp+Rel}
  & Emotion MLP~\cite{joy2020speech} & 80.29 & 0.882 & 0.802 \\
  & DyadFormer~\cite{curto2021dyadformer} & 73.39 & 0.827 & 0.733 \\
  & TRACE-LF & 82.88 & 0.916 & 0.828 \\
  & \textbf{TRACE-CE} & \textbf{93.47} & \textbf{0.999} & \textbf{0.934} \\
\hline
\multirow{4}{*}{Sp+Ctx+Rel}
  & Emotion MLP~\cite{joy2020speech} & 50.23 & 0.543 & 0.334 \\
  & DyadFormer~\cite{curto2021dyadformer} & 74.67 & 0.830 & 0.746 \\
  & TRACE-LF & 83.34 & 0.913 & 0.833 \\
  & \textbf{TRACE-CE} & \textbf{83.98} & \textbf{0.915} & \textbf{0.840} \\
\hline
\end{tabular}
\label{tab:results}
\end{table}

\subsection{Conditional Embeddings Ablation} 

\noindent\textbf{Relationship Conditioning}. Table \ref{tab:rel_accuracy} shows the accuracy gain over the speech-only baseline across relationship categories for both TRACE-LF and TRACE-CE. The two architectures exhibit clearly complementary patterns. For TRACE-LF, context conditioning is beneficial across all categories, with the strongest gains for Family ($+$12.50\,pp) and Classmates ($+$6.60\,pp), consistent with interactions where situational framing tightly constrains affective norms; relationship conditioning, however, offers negligible or negative gains for TRACE-LF across all categories. TRACE-CE shows the inverse: context conditioning degrades accuracy for all relationship types, most severely for Coworkers ($-$5.14\,pp) and Classmates ($-$4.72\,pp), while relationship conditioning yields consistent and substantial gains - Friends ($+$7.36\,pp), Romantic Partners ($+$6.74\,pp), Coworkers ($+$5.14\,pp), Family ($+$5.05\,pp), and Classmates ($+$3.77\,pp). Joint conditioning (Sp$+$Ctx$+$Rel) rarely improves over the best single-signal variant for either model; for TRACE-CE, it degrades substantially for Family ($-$9.57\,pp) and Coworkers ($-$6.07\,pp), suggesting that the two conditioning signals are competing rather than complementary within the  cross-attention mechanism. Taken together, these patterns confirm that context and  relationship are each a strong signal for entrainment detection, but their benefit is  gated by the architecture's conditioning pathway.

\begin{table}[t]
\centering\scriptsize\setlength{\tabcolsep}{9pt}
\renewcommand{\arraystretch}{1.2}
\caption{Test accuracy gain (pp) over the Speech-only baseline per relationship category. $^\dag$Small test set ($n{=}12$); estimates unreliable. ``$-$'' denotes $|\Delta|{<}0.5$\,pp.}
\label{tab:rel_accuracy}
\begin{tabular}{lrr|rr}
\toprule
 & \multicolumn{2}{c|}{\textbf{TRACE-LF} ($\Delta$pp)} & \multicolumn{2}{c}{\textbf{TRACE-CE} ($\Delta$pp)} \\
\cmidrule(lr){2-3}\cmidrule(lr){4-5}
\textbf{Relationship} & \textbf{Sp+Ctx} & \textbf{Sp+Rel} & \textbf{Sp+Ctx} & \textbf{Sp+Rel} \\
\midrule
Friends              & $+3.58$           & $-$              & $-3.15$         & $\mathbf{+7.36}$  \\
Coworkers            & $+0.93$           & $-$              & $-5.14$         & $+5.14$           \\
Classmates           & $+6.60$           & $\mathbf{+3.30}$ & $-4.72$         & $+3.77$           \\
Family               & $\mathbf{+12.50}$ & $-2.13$          & $-1.86$         & $+5.05$           \\
Romantic Partners    & $+6.43$           & $-5.23$          & $-0.52$         & $+6.74$           \\
Siblings$^\dag$      & $+25.00$          & $+8.33$          & $-$             & $+16.67$          \\
Parent--Child$^\dag$ & $+25.00$          & $-8.33$          & $-$             & $-$               \\
\bottomrule
\end{tabular}
\end{table}

\noindent\textbf{Context Conditioning.}\hspace{0.4em} Table~\ref{tab:trace_gain} reports the accuracy gain over the speech-only baseline per conversational context for both models. For TRACE-LF, context conditioning is broadly beneficial, with gains across 19 of 20 contexts; the largest improvements appear in Failure \& Lessons ($+$12.71\,pp), Support Experiences ($+$9.01\,pp), Habits \& Change ($+$8.62\,pp), and Self-Defense Event ($+$8.00\,pp), reflecting scenarios where the situational description provides strong affective framing. Regret \& Apologies is the sole context where TRACE-LF degrades under context alone ($-$7.81\,pp), consistent with the high within-context affective variability in that scenario. For TRACE-CE, the same context signal degrades accuracy across nearly all contexts, with the largest losses in Isolation \& Support Needs ($-$11.63\,pp) and Relationship Withdrawal ($-$10.71\,pp); relationship conditioning (Sp$+$Rel), by contrast, yields uniform per-context gains ranging from $+$3.23\,pp (Delivering Bad News) to $+$9.46\,pp (Support Experiences). Notably, per-context gains under TRACE-LF Sp$+$Ctx and TRACE-CE Sp$+$Rel are broadly correlated, both peak at Support Experiences and Self-Defense Event, suggesting that the two conditioning signals carry similar semantic content, but each is accessible only to the architecture whose encoder matches its textual granularity.

\begin{table}[t]
\centering\scriptsize\setlength{\tabcolsep}{6pt}\renewcommand{\arraystretch}{1.2}
\caption{Test accuracy gain (pp) over the Speech-only baseline per context category. TRACE-LF benefits from context conditioning; TRACE-CE benefits from relationship conditioning. ``$-$'' denotes $|\Delta|{<}0.5$\,pp.}
\label{tab:trace_gain}
\begin{tabular}{lrr|rr}
\toprule
 & \multicolumn{2}{c|}{\textbf{TRACE-LF} ($\Delta$pp)} & \multicolumn{2}{c}{\textbf{TRACE-CE} ($\Delta$pp)} \\
\cmidrule(lr){2-3}\cmidrule(lr){4-5}
\textbf{Context} & \textbf{Sp+Ctx} & \textbf{Sp+Rel} & \textbf{Sp+Ctx} & \textbf{Sp+Rel} \\
\midrule
Challenges \& Support              & $+2.46$           & $+0.82$          & $-4.10$  & $+7.38$          \\
Communication \& Feelings          & $+6.52$           & $-5.43$          & $-1.09$  & $+6.52$          \\
Consideration \& Misunderstandings & $+7.50$           & $+3.00$          & $-1.00$  & $+7.50$          \\
Delivering Bad News                & $+5.65$           & $+3.23$          & $-$      & $+3.23$          \\
Difficult Work Relationships       & $+1.47$           & $-$              & $-$      & $+4.41$          \\
Embarrassing Stories               & $+5.88$           & $-5.10$          & $+0.78$  & $+5.88$          \\
Failure \& Lessons                 & $\mathbf{+12.71}$ & $-9.32$          & $-2.12$  & $+5.51$          \\
Habits \& Change                   & $+8.62$           & $+0.86$          & $\mathbf{+3.45}$  & $+6.90$          \\
Isolation \& Support Needs         & $+7.75$           & $+3.49$          & $-11.63$ & $+3.88$          \\
Leadership \& Decision-Making      & $+5.41$           & $-0.68$          & $-4.73$  & $+6.76$          \\
Missed Chance to Speak Up          & $+2.78$           & $-2.78$          & $-5.56$  & $+6.94$          \\
Mistakes \& Apologies              & $+5.74$           & $-$              & $-2.05$  & $+6.97$          \\
Overcoming Challenges              & $+1.52$           & $+0.76$          & $-3.79$  & $+8.33$          \\
Past Compromise                    & $+2.38$           & $-2.38$          & $-$      & $+5.95$          \\
Regret \& Apologies                & $-7.81$           & $-3.07$          & $-2.16$  & $+6.47$          \\
Relationship Withdrawal            & $+2.14$           & $-0.71$          & $-10.71$ & $+6.43$          \\
Self-Defense Event                 & $+8.00$           & $\mathbf{+7.33}$ & $-4.67$  & $+9.33$          \\
Support Experiences                & $+9.01$           & $+5.86$          & $-1.35$  & $\mathbf{+9.46}$ \\
Trust \& Decision-Making           & $+3.57$           & $-4.17$          & $-1.79$  & $+8.93$          \\
Unresolved Disagreements           & $+4.67$           & $-3.74$          & $-1.87$  & $+6.07$          \\
\bottomrule
\end{tabular}
\end{table}

\section{Conclusion and Future Work}
In this work, we introduced DyadEE, a dataset for emotional entrainment detection in dyadic speech interaction, and TRACE, a strong temporal framework that combines window-level acoustic emotion representations with contextual and relationship information. 
Our results show that modeling emotional entrainment does not depend on the speech signal alone, as both conversational context and social relationship are important for modeling appropriate affective coordination. 
Although our approach for detecting emotional entrainment performed reliably, the non-entrained condition is instantiated via a limited set of synthetic transformations, which may not fully reflect the diversity of non-entrained conversational speech.
Future work should expand DyadEE with more naturalistic and diverse non-entrained speech, and model emotional entrainment beyond binary classification, for example as a graded or time-varying phenomenon. A particularly promising direction is deploying entrainment-aware modeling in human-LLM spoken interaction: training conversational agents to adapt their affective behavior based on TRACE predictions, and validating whether the model's entrainment scores correlate with human perception of emotion entrainment in live interactions. It would also be valuable to extend DyadEE to multimodal settings, broader social contexts, and cross-cultural or multilingual scenarios.

\section{Limitations}
TRACE operates on audio-only input, excluding visual cues (facial expression, gesture) that contribute to affective coordination in face-to-face interaction.
Non-entrained dyads in DyadEE are constructed via controlled synthetic transformations; whether models trained on these conditions generalize to naturally occurring non-entrained speech such as disengaged or mismatched conversations in the wild remains an open question. The dataset is currently English-only, and the emotional norms encoded in the relationship and context embeddings may not transfer to cross-cultural or multilingual settings. Finally, binary entrainment labels do not capture the graded, time-varying nature of affective coordination; finer-grained annotations would enable richer modeling of entrainment dynamics.
\newpage
\section{Generative AI Use Disclosure}
The authors used Claude (Anthropic) to assist with grammar checking and proofreading the manuscript. OpenAI GPT-5 was used to assist with categorizing the 798 unique Seamless Interaction speaker prompts into thematic context labels, with all prompt-to-category mappings subsequently manually reviewed and corrected by the authors. These tools were not used to generate any scientific content, results, or conclusions. All authors reviewed and take full responsibility for the final content of this paper.

\bibliographystyle{IEEEtran}
\bibliography{IEEEabrv,IEEEexample}

\end{document}